%% file: C3PortraitNet.tex
\documentclass[10pt,twocolumn,letterpaper]{article}

\usepackage{wacv}
\usepackage{times}
\usepackage{epsfig}
\usepackage{graphicx}
\usepackage{amsmath}
\usepackage{amssymb}
\usepackage[dvipsnames]{xcolor}
\usepackage{multirow}

\newif\ifdraft\draftfalse

\ifdraft

\definecolor{greyblue}{rgb}{0.1,0.6,0.5}
\newcommand\Lars[1]{\textcolor{blue}{#1}}
\newcommand\yj[1]{\textcolor{ForestGreen}{#1}}
\newcommand\hj[1]{\textcolor{Sepia}{#1}}
\newcommand\nj[1]{\textcolor{red}{#1}}

\else

\newcommand\Lars[1]{#1}
\newcommand\yj[1]{#1}
\newcommand\hj[1]{#1}
\newcommand\nj[1]{#1}

\fi
  {\begin{list}{}%
          {\setlength{\leftmargin}{#1}}%
          \item[]%
  }
  {\end{list}}

\wacvfinalcopy 

\def\wacvPaperID{387} 

\ifwacvfinal\fi
\begin{document}

\title{ExtremeC3Net: Extreme Lightweight Portrait Segmentation Networks \\
using Advanced C3-modules}

\author{
Hyojin Park \\
Seoul National University\\
{\tt\small wolfrun@snu.ac.kr}
\and
Lars Lowe Sj{\"o}sund \\
 Clova AI, NAVER Corp \\
{\tt\small lars.sjosund@navercorp.com}
\and
YoungJoon Yoo \\
 Clova AI, NAVER Corp \\
{\tt\small youngjoon.yoo@navercorp.com}
\and
Jihwan Bang \\
 Search Solutions, Inc \\
{\tt\small jihwan.bang@navercorp.com}
\and
Nojun Kwak  \\
Seoul National University\\
{\tt\small nojunk@snu.ac.kr}
}
\maketitle
\ifwacvfinal\thispagestyle{empty}\fi

\begin{abstract}

\yj{Designing a lightweight and robust portrait segmentation algorithm is an important task for \nj{a} wide range of face  applications. 
However, the problem has been considered as a subset of the object segmentation problem.}
\hj{bviously, portrait segmentation has \nj{its unique requirements}}. First, \nj{because the portrait segmentation is performed in the middle of a} whole process \nj{of} many \nj{real-world applications, it} requires extremely lightweight models.
Second, there \nj{has not been any} public datasets in this domain that \nj{contain a} sufficient number \nj{of images with unbiased statistics}. 
\yj{To solve the problems, }
we \yj{introduce} \Lars{a} new extreme\yj{ly} lightweight portrait segmentation model consisting of \Lars{a} two\nj{-}branch\yj{ed} \Lars{architecture} based on the concentrated-comprehensive convolutions block.
Our method \Lars{reduces} the number of parameter\Lars{s} from $2.1 M$ to $37.7 K$ (around 98.2\% \Lars{reduction}), \nj{while maintaining the accuracy within \Lars{a} 1\% margin from the \hj{state-of-the-art portrait segmentation method.}}
\Lars{\yj{
\Lars{In our} qualitative and quantitative analysis on the EG1800 dataset}, we show that our method outperforms various existing lightweight segmentation models.}
\hj{Second, we propose a simple method to create additional portrait segmentation data \nj{which} can \nj{improve} accuracy \nj{on} the EG1800 dataset.}
\hj{Also, we analyze the \nj{bias in} public \nj{datasets by additionally annotating} race, gender, and age \nj{on our own}. 
The augmented dataset, the additional annotations and code are available in https://github.com/HYOJINPARK/ExtPortraitSeg .}
\end{abstract}

\input{Eng/Intro.tex}

\input{Eng/Related.tex}

\input{Eng/Method.tex}

\input{Eng/Exp.tex}

\input{Eng/Conclusion.tex}

{\small
\bibliographystyle{ieee}
\bibliography{egbib}
}

\newpage
\input{SuppleEng/intro.tex}
\input{SuppleEng/histo.tex}

\input{SuppleEng/img.tex}
\input{SuppleEng/flop.tex}

\end{document}

%% file: Eng/Intro.tex
\section{Introduction}
\label{sec:intro}
     
  
\begin{figure}[t]
\begin{center}
 \begin{tabular}{c}
\includegraphics[width=0.97\linewidth]{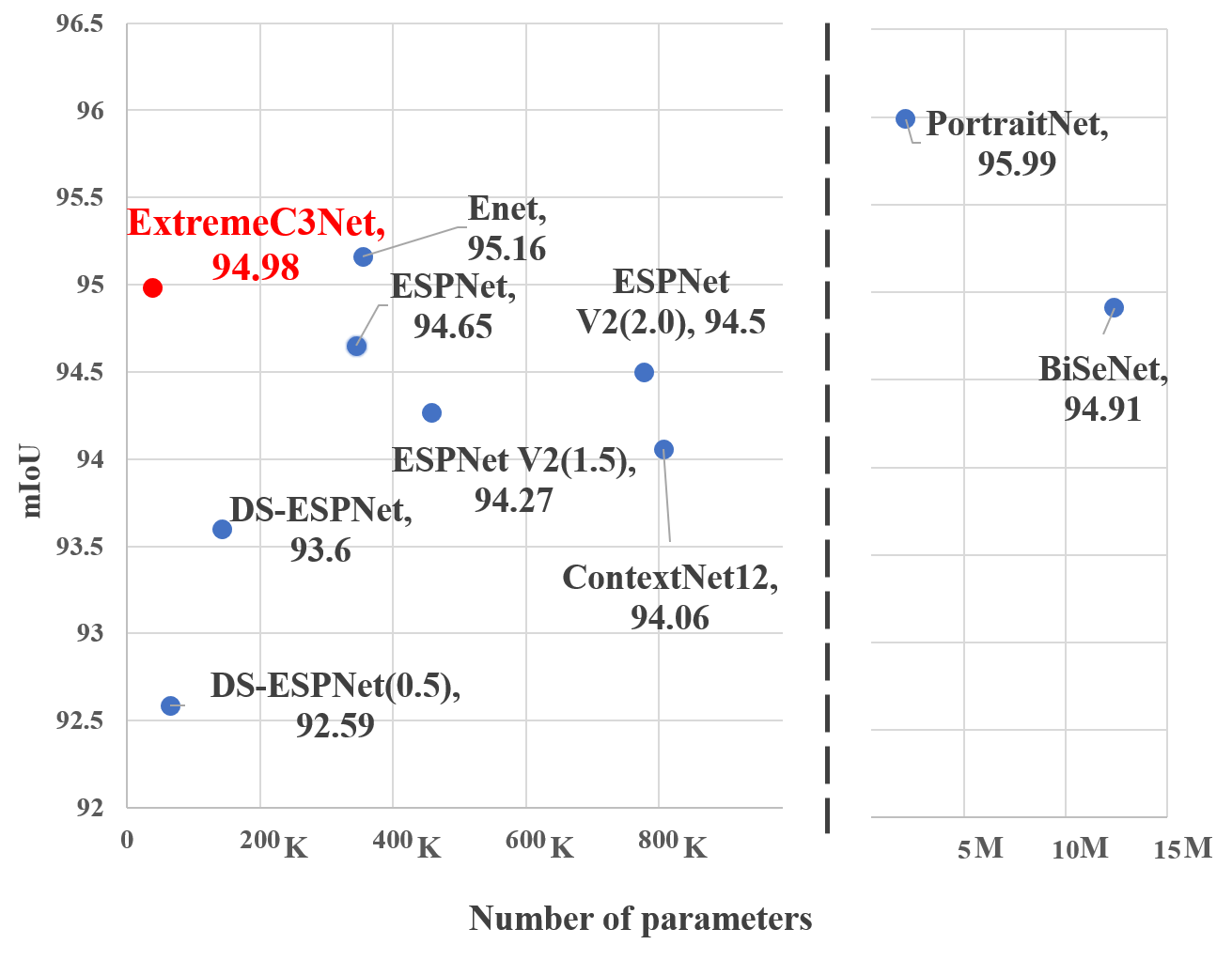}\\ 

\end{tabular}
\end{center}
   \caption{Accuracy \nj{(mIoU) vs. complexity (number of parameters) on the} EG1800 validation \nj{set.} 
   Our proposed ExtremeC3Net has high accuracy with small complexity.}
\label{fig:acc}
\end{figure}

\yj{Designing algorithms 
\Lars{working on face data} has been considered \nj{as} an important task in the computer vision field and many sub-areas such as detection, recognition, key-point extraction are actively studied.}
Among them, the portrait segmentation also has been highly used in industrial environments such as background editing, security checks, and \yj{face resolution  enhancement}~\cite{shen2016automatic, zhang2019portraitnet}.

\Lars{\hj{Because} researchers have considered the portrait segmentation problem \nj{as} a subset of semantic segmentation,
\yj{most portrait segmentation algorithms have employed general semantic segmentation algorithms~\cite{mehta2018espnet, yu2018bisenet, paszke2016enet} 
\nj{trained on a}
portrait segmentation dataset. 
\hj{
However the portrait segmentation comes with some unique obstacles.}}} 

\hj{The first thing is the small number of \nj{images in the dataset}. 
The EG1800 dataset~\cite{shen2016automatic}, a popular public portrait segmentation dataset, \nj{contains only around 1,300 training images.}
Also, we observed that this datsaet has large biases with respect to attributes such as race, age, and gender from our additional annotation.}
\yj{For general semantic segmentation tasks}, data augmentation methods like random noise, translation, and color changing are utilized to \hj{overcome the small \nj{dataset size}.}
\yj{However, it 
\Lars{is} clear that these 
methods are not sufficient for resolving the \hj{small \nj{size} and imbalance of 
portrait segmentation datasets.}}

\Lars{Second, portrait segmentation is usually \nj{used just as} one of several steps in real-world applications. Since many of these applications run on mobile devices, the segmentation model needs to be lightweight to ensure real-time speeds.}
\Lars{Researchers have developed plenty of lightweight segmentation \yj{methods}, but most of them are still not lightweight enough.}
\Lars{A few recent examples are PortraitNet~\cite{zhang2019portraitnet} with $2.1 M$ parameters, ESPNetV2~\cite{mehta2018espnetv2} with $0.78M$ parameters, and ESPNet~\cite{mehta2018espnet} with $0.36M$ parameters}. 
In general, when deploying a portrait segmentation model on a mobile device, the smaller the number of parameters without degradation of accuracy the better.
\Lars{Furthermore, one is often limited in what operations are available when deploying onto embedded systems. Therefore, it can be good to use as few different types of operations as possible.}

\yj{The contributions of the work can be summarize\nj{d} as follows:}
(1) We propose a new extremely lightweight \yj{segmentation model having less than $0.05M$ parameters as well as having competitive segmentation accuracy with PortraitNet~\cite{zhang2019portraitnet}, without using deconvolution operations.} 
\yj{Our model only uses $1.8\%$ of 
\Lars{the number of parameters of our baseline PortraitNet, 37.7K compared to 2.1M,}
but the accuracy degradation is just about $1\%$ on \Lars{the} EG1800 dataset \nj{as can be seen in Figure~\ref{fig:acc}.}} 
(2) We \yj{introduce} a simple and effective data generation method \hj{to get enough number of dataset and \yj{to alleviate the \Lars{imbalance}
of the dataset.}} 
\hj{\nj{For performance enhancement, we generated} 10,448 images for training \nj{using} our generation method.
Also, we additionally annotated the public dataset EG1800 according to race, gender, and age. \nj{In doing so,} we detected \nj{strong biases} to specific attributes.
From the experiments, we found that the proposed data augmentation can enhance the segmentation accuracy and \nj{also enrich the balance among different attributes}.
  }

%% file: Eng/Related.tex
\section{Related Work}
\label{sec:related}
%

\noindent
\textbf{Data Augmentation: } 
%
\yj{\nj{Data augmentation techniques} have become \nj{an} important factor for successful training of various deep \hj{networks}.
Many researches have shown that the augmentation alleviates over-fitting as well as enlarging the number of data samples.
In \nj{the fields of} classification and detection, not only baseline augmentation method\nj{s} such as \nj{cropping and flipping}, but also novel patch \nj{and} region-based method such as  CutOut~\cite{devries2017improved} \nj{and} CutMix~\cite{yun2019cutmix} have been proposed.}
\yj{\hj{
In semantic segmentation, basically two augmentation methods are used. 
One is image calibration methods including rotation, flip\nj{ping and} crop\nj{ping of} the images.
The other is image filtering methods controlling image attributes such as brightness, contrast \nj{and color}.
In addition to the basic augmentation methods, since the labeling cost is high, many segmentation methods additionally use crawled images with the label from relatively simple segmentation algorithms, similar to those \nj{in} text detection and localization studies~\cite{gupta2016synthetic, baek2019character}.
Jin~\etal~\cite{jin2017webly} crawled images from the web and label\nj{ed} them with dense conditional random field (CRF).
Similarly, \cite{saleh2016built,qi2016augmented, wei2016learning} applied weakly-supervised method to enlarge the datasets automatically.}}
\noindent
\textbf{Convolution Factorization: } 
Convolution factorization \yj{dividing the convolution operation into several stages \nj{has} been used} to reduce \Lars{computational complexity.}
In Inception \cite{szegedy2015going, szegedy2016rethinking, szegedy2017inception}, several convolutions are performed \Lars{in parallel, }and the results were concatenated. 
Then, a $1\times 1$ convolution \nj{is} used to reduce the number of channels. 
Xception \cite{chollet2017xception}, MobileNet \cite{howard2017mobilenets} and  MobileNetV2 \cite{sandler2018inverted} use the depth-wise separable convolution \textit{(ds-Conv)}, which performs spatial and cross-channel operations separately \Lars{to decrease computation. }
ResNeXt \cite{xie2017aggregated} and ShuffleNet \cite{zhang2017shufflenet} applied a group convolution to reduce complexity.
\yj{In segmentation, many lightweight segmentation algorithms~\cite{mehta2018espnet} also adopt the \nj{convolution factorization methods} to reduce the number of parameters.}

\noindent
\textbf{Segmentation: }
PortraitFCN+~\cite{shen2016automatic} built a portrait dataset from \hj{Flickr} and proposed \nj{a} portrait segmentation model based on FCN~\cite{long2015fully}.
After that, PortraitNet proposed \nj{a} novel portrait segmentation model with higher accuracy than PortraitFCN+ \nj{with} real-time execution time. 
Also, there are many lightweight segmentation model\nj{s}.
Enet~\cite{paszke2016enet} \Lars{was}  
the first architecture designed for real-time segmentation.
\Lars{Later, ESPNet~\cite{mehta2018espnet} improved both speed and performance by introducing an efficient spatial pyramid of dilated convolutions.}
ERFNet~\cite{romera2018erfnet} used residual connections and factorized dilated convolutions \nj{into} \hj{two asymmetric dilated convolutions}.
ContextNet~\cite{poudel2018contextnet} and FastSCNN~\cite{poudel2019fast} designed two network branches \yj{each} for global context and detail\Lars{ed} information, \nj{respectively}.
Similarly, BiSeNet~\cite{yu2018bisenet} proposed a two-paths network for preserving spatial information \yj{as well as} acquiring a large enough receptive field.
There are several works which further reduced \nj{model} parameters by combining the dilated convolution and the depth-wise separable convolution.
ESPNetV2~\cite{mehta2018espnetv2} applied \nj{the} group point-wise and depth-wise dilated separable convolutions to learn representations and showed better performance at various tasks. 
\yj{C3~\cite{park2018concentrated} resolved the degradation of accuracy from naive combination of dilated convolution and depth-wise separable convolution by proposing a concentrated-comprehensive convolution (C3).
In C3, two asymmetric convolutions are added in front of the dilated depth-wise convolution.} 
In this paper, we propose an advanced version of the C3 module which has \nj{a} much smaller \nj{model size} to satisfy the aforementioned requirement of the portrait segmentation.


%% file: Eng/Method.tex
\begin{figure*}[t]
\begin{center}
 \begin{tabular}{c c}
\includegraphics[width=0.67\linewidth]{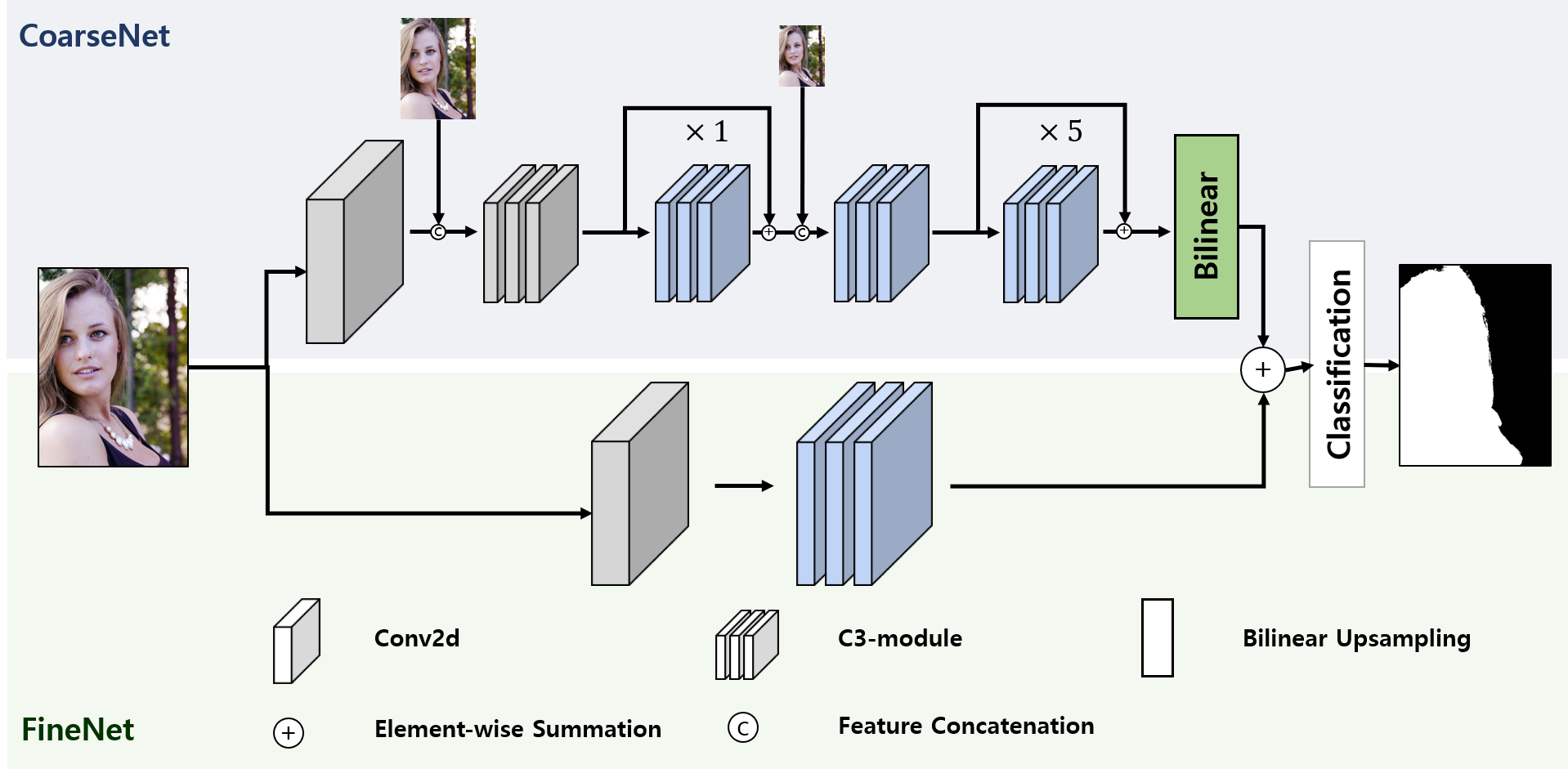} &
\includegraphics[width=0.27\linewidth]{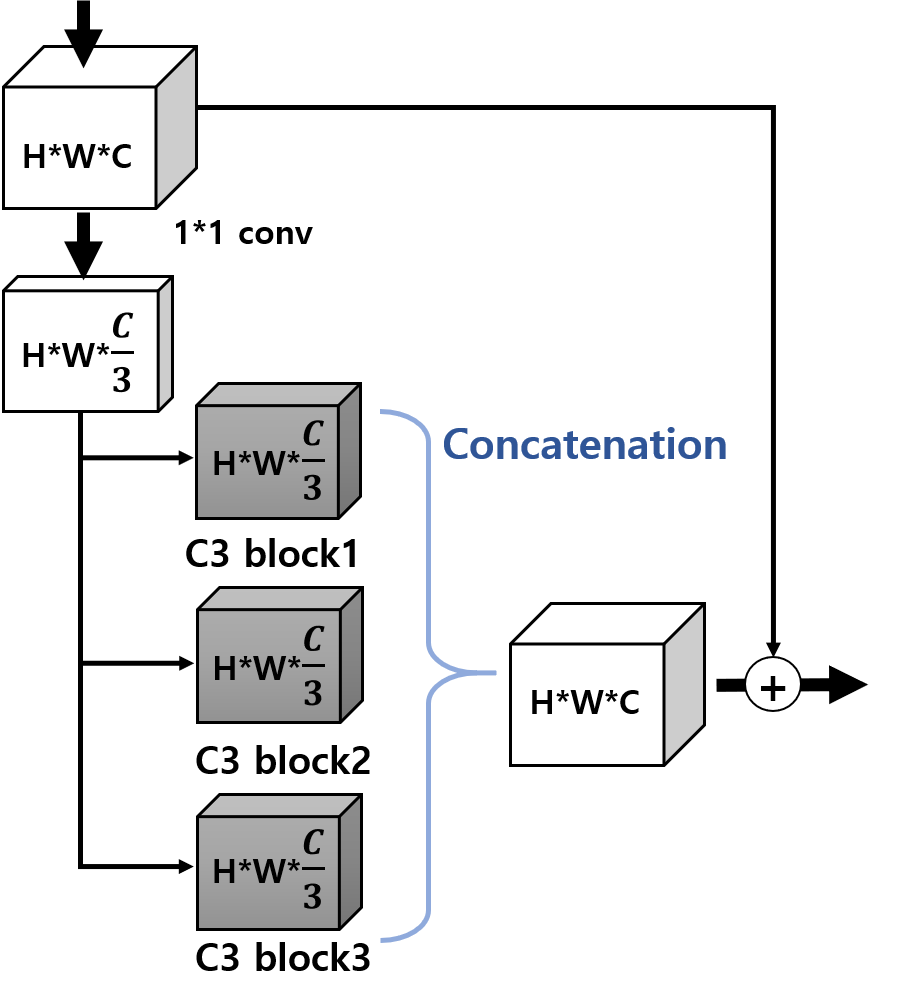}
\\(a) ExtremeC3Net &(b) Advanced C3-module
\end{tabular}
\end{center}
   \caption{(a) The model structure of ExtremeC3Net. Gray (green) color represents downsampling (unpsampling) (b) The structure of advanced C3-module consisting of three C3-blocks. }
\label{fig:structure}
\end{figure*}

\section{Method}
\label{sec:method}

In this section, \nj{w}e explain \nj{a} simple data generation framework to solve the lack of dataset in two situations.
Also, we \nj{introduce} our \hj{advanced} C3-module with \nj{a} well-designed \nj{combination of dilation ratios} and \nj{propose an} extremely lightweight segmentation model \nj{{based on it}}.

\subsection{Advanced C3-module}
\label{sec:c3}


\begin{table}[t]
  \begin{center}
 \small
 \setlength\tabcolsep{1.5pt}
    \begin{tabular}{ l|c| ccc | l| c | ccc }
    \hline
         & Input  & B1 & B2& B3 &      & Input  & B1 & B2 & B3 \\
           \hline \hline
    L1    & $112\times112\times27$ & 1     & 2     & 3     & L5    & $56\times56\times56$ & 2     & 4     & 8 \\
    L2    & $56\times56\times48$ & 1     & 3     & 4     & L6    & $56\times56\times56$ & 2     & 4     & 8 \\
    L3    & $56\times56\times99$ & 1     & 3     & 5     & L7    & $56\times56\times56$ & 2     & 4     & 8 \\
    L4    & $56\times56\times56$ & 2     & 4     & 8     & L8    & $56\times56\times56$ & 2     & 4     & 8 \\
     \hline
    \end{tabular}%
    \end{center}
     \caption{Detail\Lars{ed} setting of \Lars{the dilation} ratio. L denotes layer and B denotes block in each C3-module. }
  \label{tab:dilate}%
\end{table}%

\Lars{Stacking multiple dilated convolutional layers increases the receptive field and has been used to boost performance in many segmentation models.}
For further reducing model complexity, \nj{depth-wise separable convolutions are} widely used.
\Lars{ \hj{However, concentrated-comprehensive convolutions block (C3)~\cite{park2018concentrated} pointed out that \Lars{the} over-simplified operation \Lars{performed} by \Lars{the} depth-wise separable dilated convolution  causes  severe  performance degradation due to loss of information contained in the feature map\Lars{s}.}}
When we combine a dilated convolution with depth-wise separable convolution for our basic module, we also observe the same degradation.
\Lars{To mitigate this problem \cite{park2018concentrated} designed the C3-block, \hj{which is \nj{composed} of a concentrat\nj{ion} stage and a comprehensive convolution stage.
The concentrat\nj{ion} stage compresses information from the neighboring pixels to alleviate the information loss.}}
\Lars{It does} not use \Lars{the} standard square depth-wise convolution but 
\Lars{instead uses an} asymmetric depth-wise convolution for reducing complexity.

\Lars{In this paper, we use the C3-module, but \nj{unlike in \cite{park2018concentrated}, we} reduce the number of C3-blocks from 4 to 3, as shown in Figure \ref{fig:structure}(b).} 
We also re-designed the dilation ratio 
\Lars{for each C3-module, based on insights about neural network kernel properties from Zeiler \etal~\cite{zeiler2014visualizing}}.
A kernel which is close to the input image extracts common and local features,
\Lars{while one close to the classifier extracts more class-specific and global features.}
The \nj{models} C3 \cite{park2018concentrated} and ESPNet \cite{mehta2018espnet} \Lars{use the same dilation ratio combination for every module in their architectures, and \nj{do not} take this kernel property into account. }
\Lars{We design these ratios more carefully, using small ratios in modules close to the input, and larger ratios for later modules.}
Table~\ref{tab:dilate} shows the detailed \Lars{dilation ratio settings for each layer.}

\begin{figure*}[t]
\begin{center}
 \begin{tabular}{c c}
\includegraphics[width=0.45\linewidth]{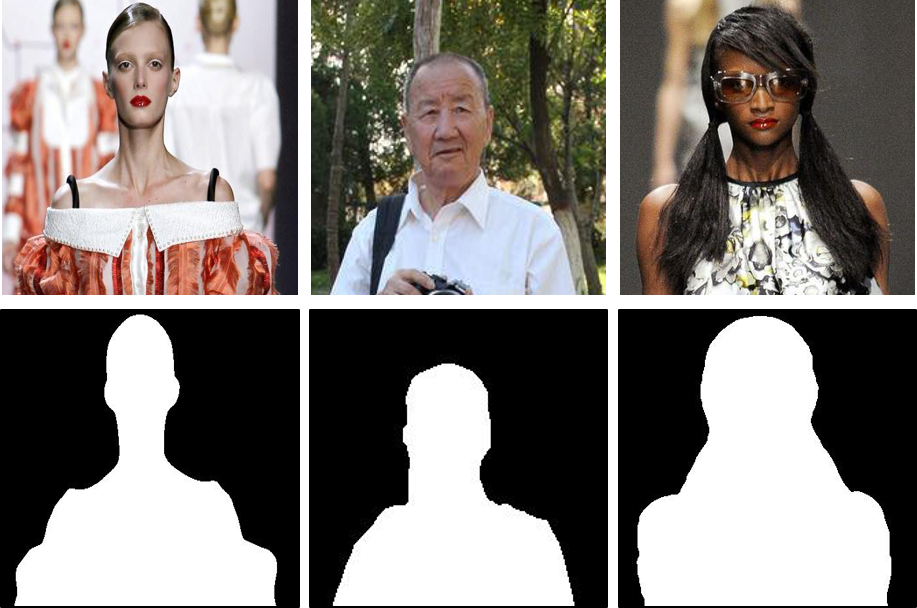} &
\includegraphics[width=0.45\linewidth]{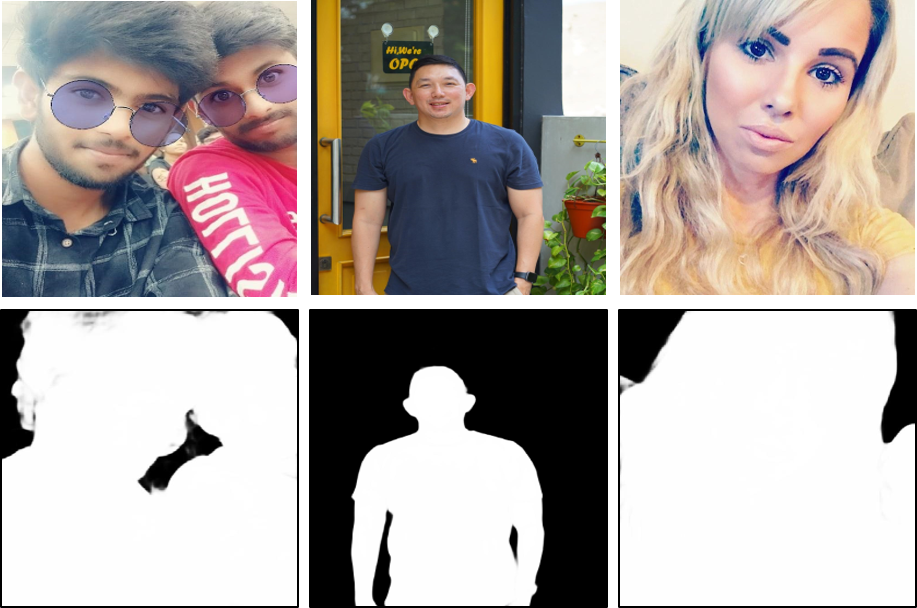}
\\(a) The situation of having human segmentation ground truths &(b) The situation of having only raw images
\end{tabular}
\end{center}
   \caption{Examples of \Lars{images and segmentation masks} generated by our proposed framework in two situations (a) \Lars{Data generated from the Baidu human segmentation dataset, by using a face detector. (b) Data generated from raw images by using a segmentation model.}}
\label{fig:Gdata}
\end{figure*}

\subsection{ExtremeC3Net Architecture}
\label{sec:arch}


In this \nj{part}, we introduce \nj{the architecture of} our ExtremeC3Net \nj{for} segmentation.  We use a two-branched architecture \Lars{for higher model efficiency, similar to BiseNet~\cite{yu2018bisenet}, ContextNet~\cite{poudel2018contextnet}, and Fast-SCNN~\cite{poudel2019fast}. }
\Lars{\nj{The authors of} \cite{poudel2018contextnet} and \cite{poudel2019fast} argue that a two-branch architecture has different properties compared to the common encoder-decoder structure.}

The encoder-decoder structure extracts feature\Lars{s using an} encoder network and recovers \Lars{the} original resolution \Lars{using a} decoder network, \Lars{which often uses deconvolution operations and concatenation with the encoder feature maps.} 
The amount of computation of \Lars{the} decoder is increased proportionally to the resolution \nj{reduced in} the encoder.

The two-branch architecture is composed of a deep network and a shallow network \Lars{running in parallel}.
The deep network learns complex and global context feature\Lars{s} and \Lars{the} shallow network preserves spatial detail.
\Lars{Due to this preservation of detail, there is less need to have a sophisticated decoder module, as we can use the simpler bilinear upsampling operation for resolution recovery. This removes the need to use the deconvolution operation, and reduces the number of parameters.}

\nj{As shown in Figure~\ref{fig:structure}, we} design \Lars{a} two-branched architecture consisting of \Lars{a} CoarseNet\Lars{-branch} and \Lars{a} FineNet\Lars{-branch}. 
The CoarseNet\Lars{-branch} extracts deep feature embedding\Lars{s} and \Lars{the} FineNet\Lars{-branch} has responsibility for preserving spatial detail.
Finally, the feature maps from \Lars{the two branches are combined using element-wise addition.} 

The CoarseNet\Lars{-branch} is constructed from a series of advanced C3-module\Lars{s} and gives a rough segmentation. 
It reduces \nj{the size of the feature map} to \nj{a quarter} by \Lars{first applying} a \nj{convolution with stride 2} and an advanced C3-module.
\hj{After th\Lars{at}, seven C3-modules sequentially produce feature maps without \nj{the} downsampling operation.}
Each C3-module has a different combination of \nj{dilation} ratio\Lars{s}, as mentioned above, for making \nj{better features}.
\Lars{We concatenate the downsampled input image onto the feature maps at two different points in the branch. ESPNet showed that this practice can increase the information flow and accuracy.}

\Lars{The FineNet-branch creates more exact boundary lines by only downsampling the feature maps by a factor 2. This branch is kept shallow, as the large feature size would cause the computational complexity to grow rapidly with network depth.}
First, it reduces \Lars{the} feature map \Lars{size} by \Lars{applying a convolution with stride 2. Then it applies a C3-module to capture spatial detail information.} 

A point-wise convolution is applied to the last feature \Lars{maps from CoarseNet, to get the same number of channels as from FineNet. }
Bilinear upsampling increases \Lars{the output from CoarseNet and FineNet with a factor 4 and 2 respectively. }
\Lars{The two} full resolution feature maps are \Lars{then} aggregated by element-wise summation.

The intersection over union(IoU) score is widely used to evaluate segmentation quality\Lars{, }
\Lars{and can be directly optimized using the Lov\'{a}sz-Softmax loss~\cite{berman2018lovasz} as a surrogate function.}
\Lars{We apply this loss function both as the main segmentation loss and also as an auxiliary loss focused around the boundary area.}
\Lars{We define the boundary area as the non-zero part of the difference between the morphological dilation and erosion of the binary ground truth segmentation mask.}
The final loss is as shown in Equation~\ref{eq:loss}.
Lov\'{a}sz denotes a Lov\'{a}sz-Softmax loss and $f$ is a $7\times7$ size filter \Lars{used} for \Lars{the} dilation and erosion operation\Lars{s}.
\hj{$\mathcal{P}$ denotes the ground truth, and $\mathcal{B}$ is \Lars{the} boundary area 
\Lars{as} defined by \Lars{the} morphology operation. 
The $i(j)$ is an included pixel in \Lars{the} ground truth(boundary area).}
$y^*$ is a binary ground truth value and $\hat{y}$ is a  predicted label from a segmentation model.

\begin{equation}
\begin{aligned}
& \mathcal{B} = (f \oplus mask) - (f \ominus mask) \\
& Loss = \text{Lov\'{a}sz}_{i \in \mathcal{P}}(y_i^*, \hat{y}) + w\text{Lov\'{a}sz}_{j \in \mathcal{B}}(y_j^*, \hat{y}_j)
\end{aligned}
\label{eq:loss}
\end{equation}



\subsection{Data Generation Method}
\label{sec:data}

\begin{table*}[t]
  \begin{center}
\begin{tabular*}{0.99\textwidth}{@{\extracolsep{\fill}}| l |  c cccc|}
\hline
    Method & Parameters & FLOPs(all) (G) & FLOPs~\cite{mehta2018espnetv2} (G) & mIoU  & Paper~\cite{zhang2019portraitnet} \\
    \hline
    \hline
  Enet (2016) \cite{paszke2016enet}  & {355 K} &  {0.703} &  {0.346 } & 95.16 & 96.00 \\
   BiSeNet (2018) \cite{yu2018bisenet} &  {12.4 M} &  {4.64} &  {2.31 } & 94.91 & 95.25 \\
    PortraitNet (2019)\cite{zhang2019portraitnet}  &{2.08 M} &  {0.666 } &  {0.325} & 95.99 & 96.62 \\
    ESPNet (2018)\cite{mehta2018espnet} & 345 K &  0.665 & 0.328  & 94.65 &  \\
   DS-ESPNet  &143 K & 0.418 & 0.199 & 93.60  &  \\
    DS-ESPNet(0.5)  & 63.9K & 0.296 & 0.139 & {92.59} &  \\
   ESPNetV2(2.0) (2019)\cite{mehta2018espnetv2}  & {778 K} &  {0.476} &  {0.231} & 94.50  &  \\
  ESPNetV2(1.5) (2019)\cite{mehta2018espnetv2}  &  {458 K} &  {0.285} &  {0.137} & 94.27 &  \\
   ContextNet12 (2018)\cite{poudel2018contextnet} & 838 K & 3.17 & 1.55 & 95.71 &  \\
    ContextNet12(0.25) &   67.2 K & 0.372 & 0.176  &   93.24    &  \\
    \textbf{ExtremeC3Net(Ours)}  & \textbf{37.7 K} & \textbf{0.286} & \textbf{0.128} &\textbf{94.23} &  \\
   \textbf{ExtremeC3Net(Ours + generated dataset)}  & \textbf{37.7 K} & \textbf{0.286} & \textbf{0.128} &\textbf{94.98} &  \\
    \hline
    \end{tabular*}%
    \end{center}
     \caption{EG1800 validation results for the proposed ExtremeC3Net and other segmentation models. DS denotes depth-wise separable convolution. FLOPs(all) counted all the operation FLOPs, and FLOPs~\cite{mehta2018espnetv2} calculated the number according to ESPNetV2~\cite{mehta2018espnetv2} official code. Performances of $6_{th}$ column were reported in the PortraitNet\cite{zhang2019portraitnet} paper. }
  \label{tab:exp}%
\end{table*}%


\Lars{Annotating data often comes with high costs, and the annotation time per instance varies a lot depending on the task type. For example, Papadopoulos \etal\cite{bearman2016s} estimate the annotation time per instance for PASCAL VOC to be 20.0 seconds for image classification and 239.7 seconds for segmentation, an order of magnitude difference.}
To mitigate \nj{the} cost of annotation \Lars{for portrait segmentation}, we consider \nj{a couple of plausible situations:} 1) having \nj{images with ground truth human segmentation} 
2) having only raw images.
We \nj{make use of either an} elaborate face detector model \nj{(case 1)} or \nj{a} segmentation model \nj{(case 2)} for \nj{generating} pseudo ground \nj{truths} to each situation. 

When we have human images and ground truths, we just need a bounding box around \Lars{the} portrait area.
We took plenty of images from Baidu dataset~\cite{wu2014early}\Lars{, which contains }
5,382 human full body segmentation \Lars{images} covering various poses, fashions and backgrounds.  
\yj{To get the bounding box and portrait area, we detect the face location of the images using \Lars{a} face detector~\cite{yoo2019extd}.}
\yj{Since the face detector tightly bounds the face region,} we \Lars{increase the bounding box size to include parts of the upper body and background before cropping the image and ground truth segmentation.}

\Lars{We also create a second dataset from portrait images scraped from the web, applying a more heavyweight segmentation model to generate pseudo ground truth segmentation masks.}
\Lars{This segmentation model consists of a DeepLabv3+~\hj{\cite{deeplabv3plus2018}} architecture with a SE-ResNeXt-50~\hj{\cite{xie2017aggregated}} backbone. The model is pre-trained on ImageNet and finetuned on a proprietary dataset containing around 2,500 fine grained human segmentation images. The model is trained for general human segmentation rather than for the specific purpose of portrait segmentation. Despite this the model works well for the portrait segmentation task, and can be used for acquiring extra training data. 
}
 
Finally, human annotators just check \Lars{the quality of each pseudo ground truth image, removing obvious failure cases. }
This method \Lars{reduces} the annotation effort per instance from \Lars{several minutes }
to 1.25 seconds \Lars{by transforming the segmentation task into a binary classification task. Examples of the generated dataset are shown in Figure~\ref{fig:Gdata}.}

%% file: Eng/Exp.tex
\section{Experiment}
\label{sec:exp}
\begin{figure*}[t]
\begin{center}
\begin{tabular}{c c}
    \includegraphics[width=0.48\linewidth]{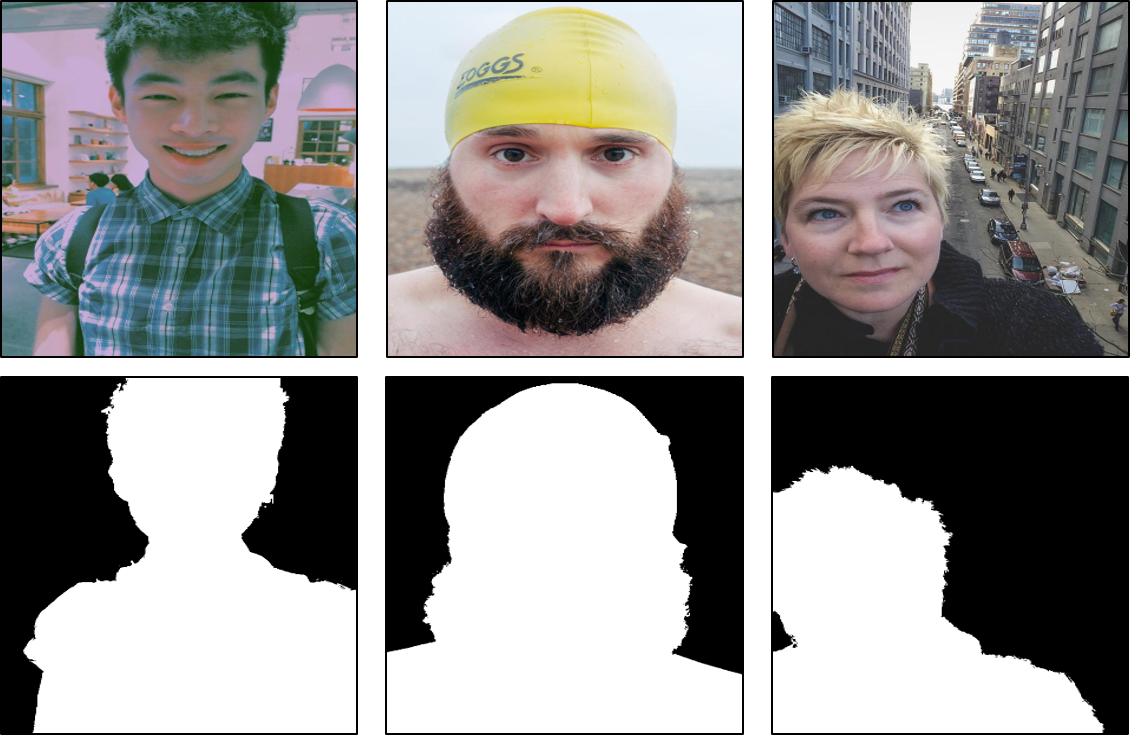} & 
     \includegraphics[width=0.48\linewidth]{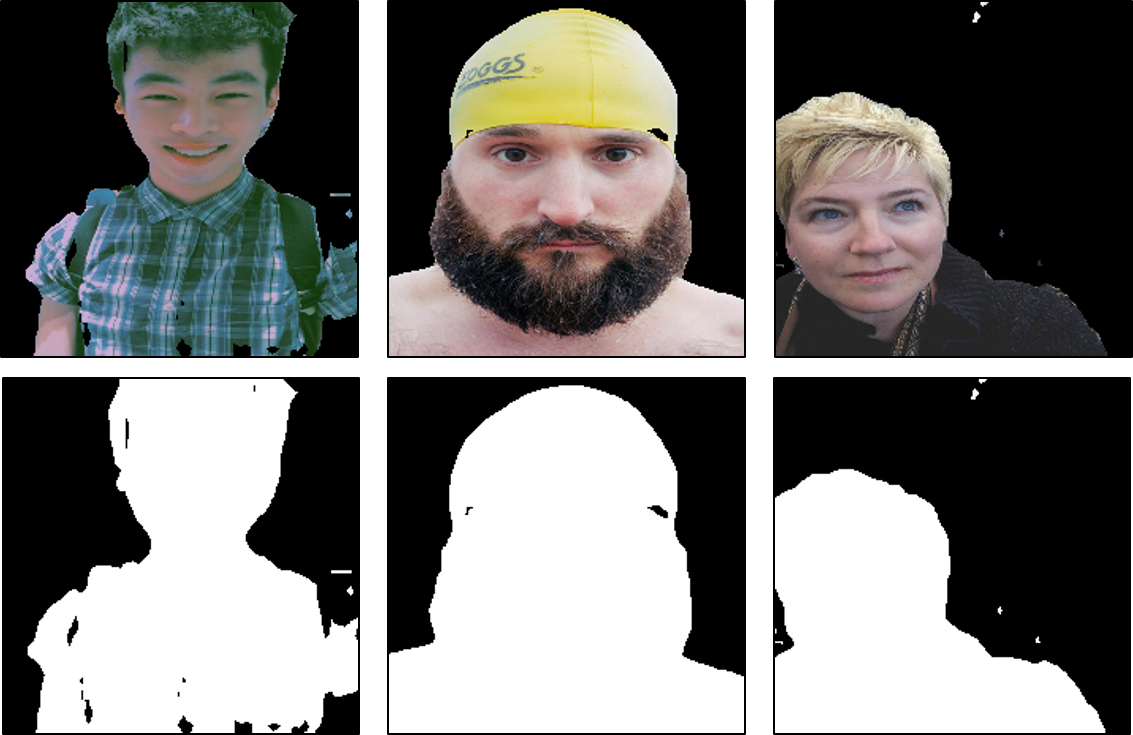} \\
    (a) Input images and ground-truths & \textbf{(b) Ours ExtremeC3Net param : 37.7 K, mIoU : 94.98 }  \\
     \includegraphics[width=0.48\linewidth]{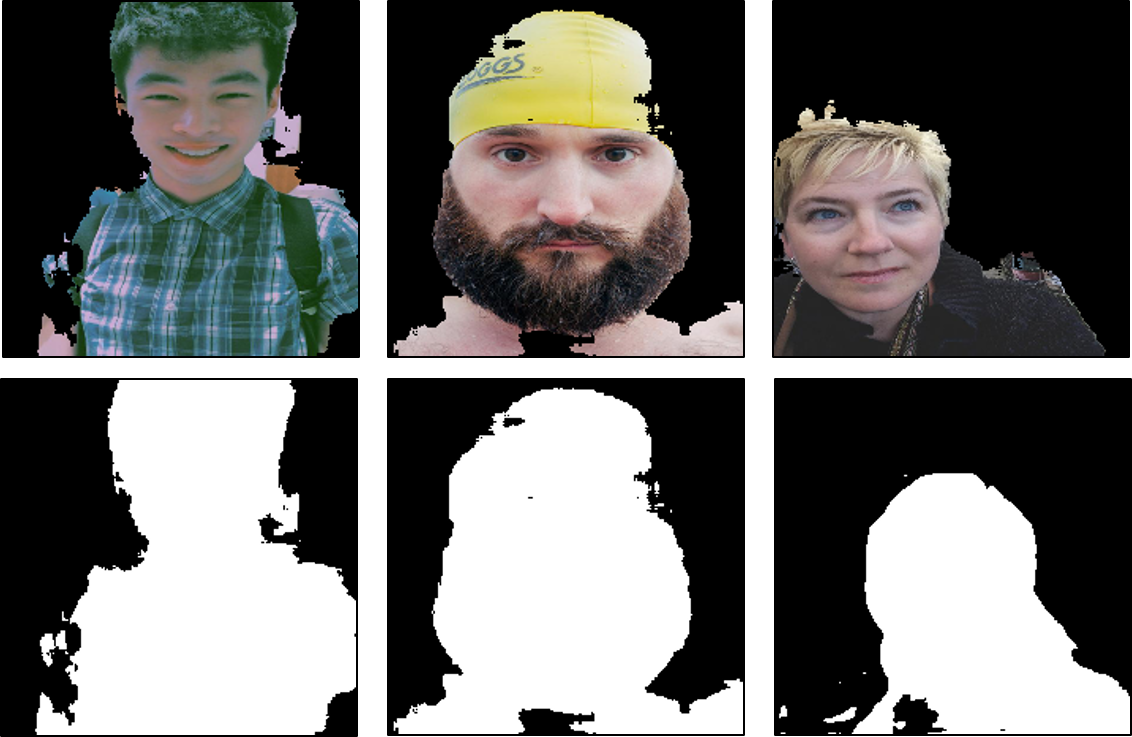}
     & \includegraphics[width=0.48\linewidth]{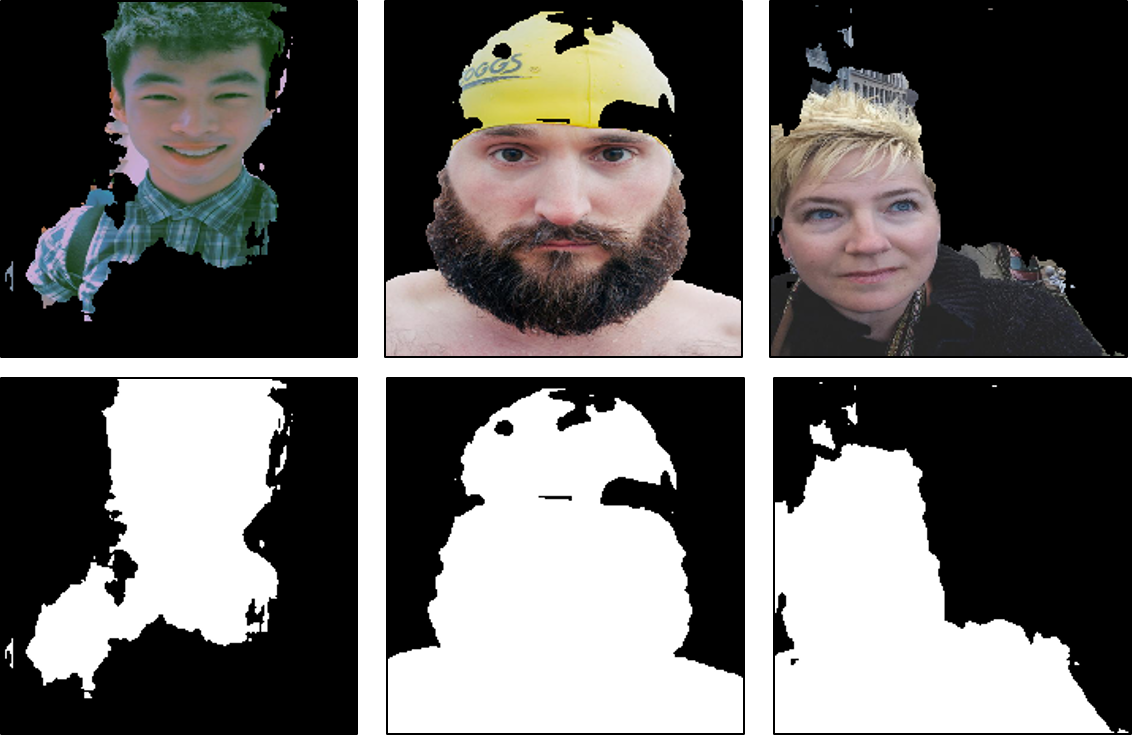} \\
    (c) DS-ESPNet(0.5) Param : 63.9 K, mIoU : 92.59  & (d) ContextNet12(0.25) Param : 67.2 K, mIoU : 93.24 \\
        \end{tabular}%
\end{center}
   \caption{Qualitative comparison results on the EG1800 validation dataset.}
\label{fig:ex}
\end{figure*}

We evaluated the proposed method \yj{on the} public dataset EG1800~\cite{shen2016automatic}, which collected images from Flickr with \yj{manually} annotated \yj{labels}.
\yj{The dataset has a total of $1,800$ images and is divided into $1,500$ train and $300$ validation images}.
However, we could access only 1,309 images for train and 270 for validation, since some of the URLs are broken. 
We built additional 10,448 images \yj{from the proposed} data generation method \yj{mentioned in Section ~\ref{sec:data}.}

We trained our model using ADAM optimizer \yj{with initial learning rate to $1e^{-3}$, batch-size to $60$,  weight decay to $5e^{-4}$.}
\yj{We trained the model with total 600 epochs, and image resolution was set to $224 \times 224$.}
\yj{We used a two-stage training method for training the proposed method.
For the first 300 epochs, we only trained the \hj{CoarseNet-branch}.
Then, with intializing the the best parameters of the \hj{CoarseNet-branch} from the previous step, we trained the overall ExtremeC3Net model \Lars{for an additional 300 epochs. }}
We evaluated our model \yj{followed by various ablations}  using mean intersection over union (mIoU) and compared with a SOTA portrait segmentation model\yj{ including} other lightweight segmentation models.

Finally, we demonstrated the results of our additional annotation about detailed attributes of faces.
The dataset covers many different images, but \yj{the analysis below shows that the dataset is biased to the specific races or ages.}
\yj{From Figure~\ref{fig:att_histo}, we can see that Caucasian occupies majority portion in the dataset. 
Also, the age group is mainly biased to `Youth and Middle-aged' group.}
We give a guideline which attribute is more critical to segmentation accuracy from our analysis and accuracy improvement from our data generation method.


\subsection{Evaluation Results on the EG1800 Dataset}
We compared \yj{the} proposed model \yj{to} PortraitNet\cite{zhang2019portraitnet}, which has SOTA accuracy in the portrait segmentation \yj{field}. 
\yj{Since some sample URLs in the EG1800 dataset are missing}, we re-trained the PortraitNet followed the original method in paper and official code \yj{from the remaining samples in EG1800 dataset}.
The PortriatNet compared their work \yj{to} BiseNet, Enet, and PortraitNet. 
\yj{Therefore}, we also re-trained BiSeNet and ENet follow\yj{ing} the method of PortraitNet for a fair comparison.
\yj{As shown in Table \ref{tab:exp}, }the accuracy of the re-trained results are slightly decreased due to the reduced size of the training dataset.
\yj{The aforementioned approaches only} counted floating-point operations (FLOPs) of convolution and batch normalization layers, which occupy a large portion of \yj{the total} FLOPs.
However, other operations such as activations and deconvolutions also affect total FLOPs, and these operations are hard to ignore \yj{when it comes to the} lightweight model \yj{case}.
\yj{Therefore, in addition to the FLOPs counting in the official ESPNetV2 code, we also measured the FLOPS including all the operations. The calculation method is described in Supplementary material.}

\yj{Among the comparison methods}, DS-ESPNet has the same structure with ESPNet, \yj{with only changing the standard dilated convolutions of the model} into depth-wise separable dilated convolutions.
For ESPNetV2~(2.0) and ESPNetV2~(1.5), \yj{we} changed the channel of convolution kernels to reduce the model size. 
\yj{We also reduced the channel of convolution kernels in DS-ESPNet~(0.5) and ContextNet12~(0.25)} by half and quarter from the original models \yj{to make the model less} than 100K parameters and 0.2G FLOPs.

\yj{From Table~\ref{tab:exp}, we can see that} our proposed method showed comparable or better performance than \yj{the other models with less number of parameters and FLOPs.}
The SOTA PortraitNet \yj{showed} the highest accuracy in all the experimental results, and \yj{has achieved even} better performance than the heavier BiSeNet.
However, still, PortraitNet requires a large number of parameters, \yj{which is a disadvantageous for using it on smaller devices.}
\yj{The proposed} ExtremeC3Net has reduced the number of parameters by 98\%, and FLOPs is reduced by half compared to PortraitNet, while maintaining accuracy.
ESPNet and ESPNet V2 have similar accuracy, 
but showed a trade-off between the number of parameters and FLOPs.
ESPNet V2 has more 
parameters than ESPNet, but ESPNet needs more FLOPs than ESPNet V2.
Enet shows better performance than both models but requires more FLOPs.
In comparison, our proposed method has less number of parameters and FLOPs, but better accuracy than ESPNet and ESPNet V2.
Our model accuracy is 0.18\% less than Enet, but our number of parameters is around 10 times less than Enet.
In particular, our ExtremeC3Net has the highest accuracy in an extremely lightweight environment.
Both DS-ESPNet(0.5) and ContextNet12(0.25) \Lars{have a larger} \yj{number of parameters}, but their scores \yj{were} much lower than our method.
Figure \ref{fig:ex} \Lars{shows that the quality of our model is superior to other extremely lightweight models.}

We compared \yj{the execution speed of the proposed model} with SOTA portrait segmentation model PortraitNet on an Intel Core i7-9700 CPU environment with the PyTorch framework.
PortraitNet needs 0.119 sec, but ExtremeC3Net takes only 0.062 sec for processing an image.
\yj{In summary,} \yj{the proposed} ExtremeC3Net showed outstanding performance among the various segmentation model in terms of accuracy and speed.

\begin{table}[t]
 \begin{center}
    \begin{tabular}{|l|c|}
    \hline
    Method & mIoU \\
    \hline\hline
      Baseline with advanced C3-module & 94.09 \\
     + Using generated pseudo dataset  & 94.27 \\
     + Changing cross-entropy loss into Lov\'{a}sz loss   & 94.45 \\
     + Adding auxiliary loss into boundary area  & 94.97 \\
    \hline
     + Applying our methods without pseudo dataset & 94.23 \\
    \hline
    \end{tabular}%
    \end{center}
\caption{Ablation study results about our proposed method.}
  \label{tab:ablation}%
\end{table}%

\begin{table}[t]
 \begin{center}

    \begin{tabular}{| l l | c|}
    \hline
    Method & Position of large ratios & mIoU \\
    \hline\hline
    Baseline setting & All 8 layers   & 93.16 \\
    Advanced setting  & L4, L5, L6, L7, L8 & 94.09 \\
    Reverse setting & L1, L2, L3 & 92.33 \\
    \hline
     \end{tabular}%
    \end{center}
      \caption{ Ablation study results about adjusting the combination of dilated ratios in the C3-modules.}
  \label{tab:d-setting}%
\end{table}%

\subsection{Ablation Studies}

Table \ref{tab:ablation} shows the accuracy improvement from \yj{the various ablations.}
The baseline model is trained with cross-entropy loss \Lars{using the} EG1800 dataset.
\Lars{Applying Lov\'{a}sz loss, the auxiliary loss, and increasing dataset size all enhance the mIoU. }
When we used only the Lov\'{a}sz loss and auxiliary loss without \Lars{the additional }10,448 images, the accuracy was improved slightly.
However, when we simultaneously applied our loss methods and \Lars{the large additional dataset }
the accuracy was increased from $94.23$ to $94.97$.

Table \ref{tab:d-setting} shows the importance of the \yj{combination of dilation ratios of the convolutional filters} in the extremely lightweight model 
\yj{depending on the position of the filter.}
Our ExtremeC3Net \yj{consists of} eight advanced C3-modules.
The baseline setting \yj{uses} the same dilation ratios $d$ in all 3C-modules, $d=[2,4,8]$.
The advanced setting \yj{uses} small dilation ratios \Lars{at} \yj{the} shallow layers, which are close to \Lars{the} input image, and \yj{uses} large ratios \Lars{at} \yj{the} deeper layers, which are far from \Lars{the} input image, \yj{as \Lars{shown} in Table \ref{tab:dilate}.}
The reverse setting \yj{applied the dilated ratio of the filters} by the opposite direction, which means small\yj{er} dilated ratios located in deeper layers.  
Our advanced setting performed better than baseline setting, and reverse setting showed much lower accuracy than \yj{the} others.
From the results, we can see that \yj{making small receptive fields for the local features} in the shallow layers is critical \Lars{for} extremely lightweight portrait segmentation.

\begin{table}[h]
  \begin{center}
    \begin{tabular}{|c|l| cccc |}
    \hline
          \multicolumn{2}{|c|}{}        & \# Img & Exp1 & Exp2 & Ours\\
          \hline \hline
    \multirow{3}[0]{*}{Race} & Caucasian & 214   & 93.67  & 94.51 & \textbf{94.81}  \\
          & Asian & 46    & 94.31  & \textbf{94.78}  & 93.79  \\
          & Black & 11    & 94.74  & 95.05  & \textbf{95.84}  \\
          \hline
    \multirow{2}[0]{*}{Gender} & Man   & 112   & 92.85  & 94.21  & \textbf{94.43}\\
          & Woman & 159   & 94.54  & 94.81  & \textbf{94.91} \\
          \hline
    \multirow{3}[0]{*}{Age} & Child & 42    & 94.47  &93.86& \textbf{95.10}   \\
          & Youth & 219   & 93.87  &\textbf{94.79} & 94.63  \\
          & Senior & 10    & 90.07  & 92.97 & \textbf{93.98}  \\
          \hline
       \multicolumn{2}{|c|}{Accuracy} &  271  & 94.09     & 94.59     & 94.98 \\
          \hline
    \end{tabular}%
    \end{center}
  \label{tab:att_acc}%
  \caption{ Validation results of each attribute about different data augmentation methods and our proposed method. The reason why total number of image is not 270 but 271 that there are two people in one validation image. A class of "Child" includes infant or child and a class of "Youth" denotes youth or middle-aged person. "Exp" means experiment.}
\end{table}%

\begin{figure*}[t]
\begin{center}
\includegraphics[width=0.99\linewidth]{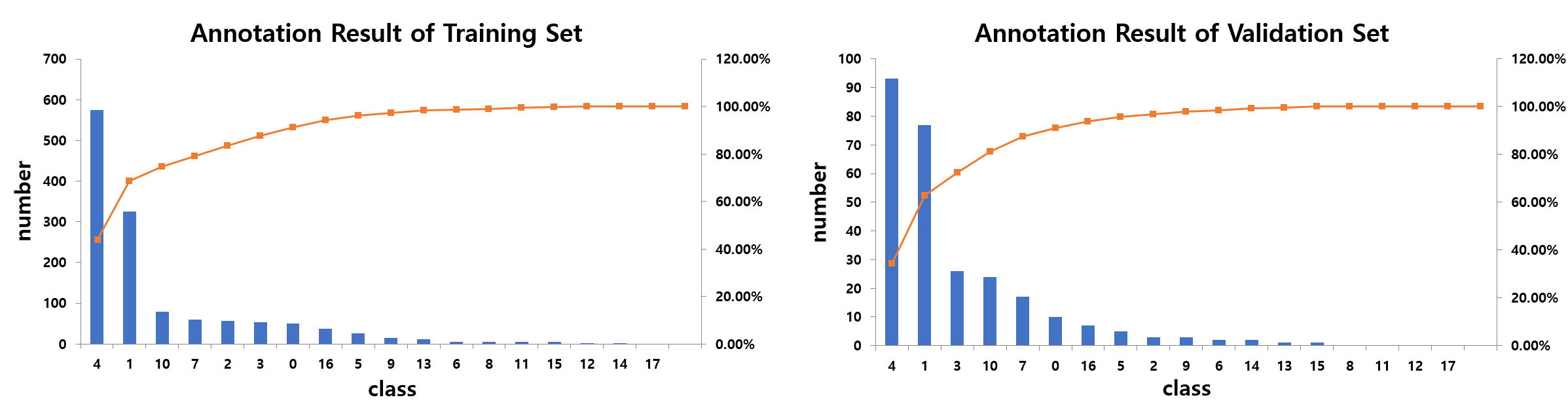}
\end{center}
   \caption{Dataset histogram about the number of each group image. The detailed number is described in supplementary material}
\label{fig:att_histo}
\end{figure*}

\begin{table*}[h]
\begin{center}

    \begin{tabular}{| c | ccc | c |ccc | c | ccc |}
    \hline
    Class & Race & Gender & Age  & Class & Race & Gender & Age  & Class & Race & Gender & Age \\
     \hline  \hline
    0     & Caucasian & Man   & Child    & 6     & Asian & Man   & Child     & 12    & Black & Man   & Child  \\
    1     & Caucasian & Man   & Youth   & 7     & Asian & Man   & Youth   & 13    & Black & Man   & Youth \\
    2     & Caucasian & Man   & Senior & 8     & Asian & Man   & Senior & 14    & Black & Man   & Senior \\
    3     & Caucasian & Woman & Child    & 9     & Asian & Woman & Child     & 15    & Black & Woman & Child  \\
    4     & Caucasian & Woman & Youth    & 10    & Asian & Woman &Youth    & 16    & Black & Woman & Youth \\
    5     & Caucasian & Woman & Senior & 11    & Asian & Woman & Senior & 17    & Black & Woman &Senior  \\
      \hline
    \end{tabular}%
      \end{center}
     \caption{Classes of attributes. "Child" includes infant or child and "Youth" denotes youth or middle-aged person.}
   
  \label{tab:att_label}%
\end{table*}%

\begin{figure}[t]
\begin{center}
\includegraphics[width=0.99\linewidth]{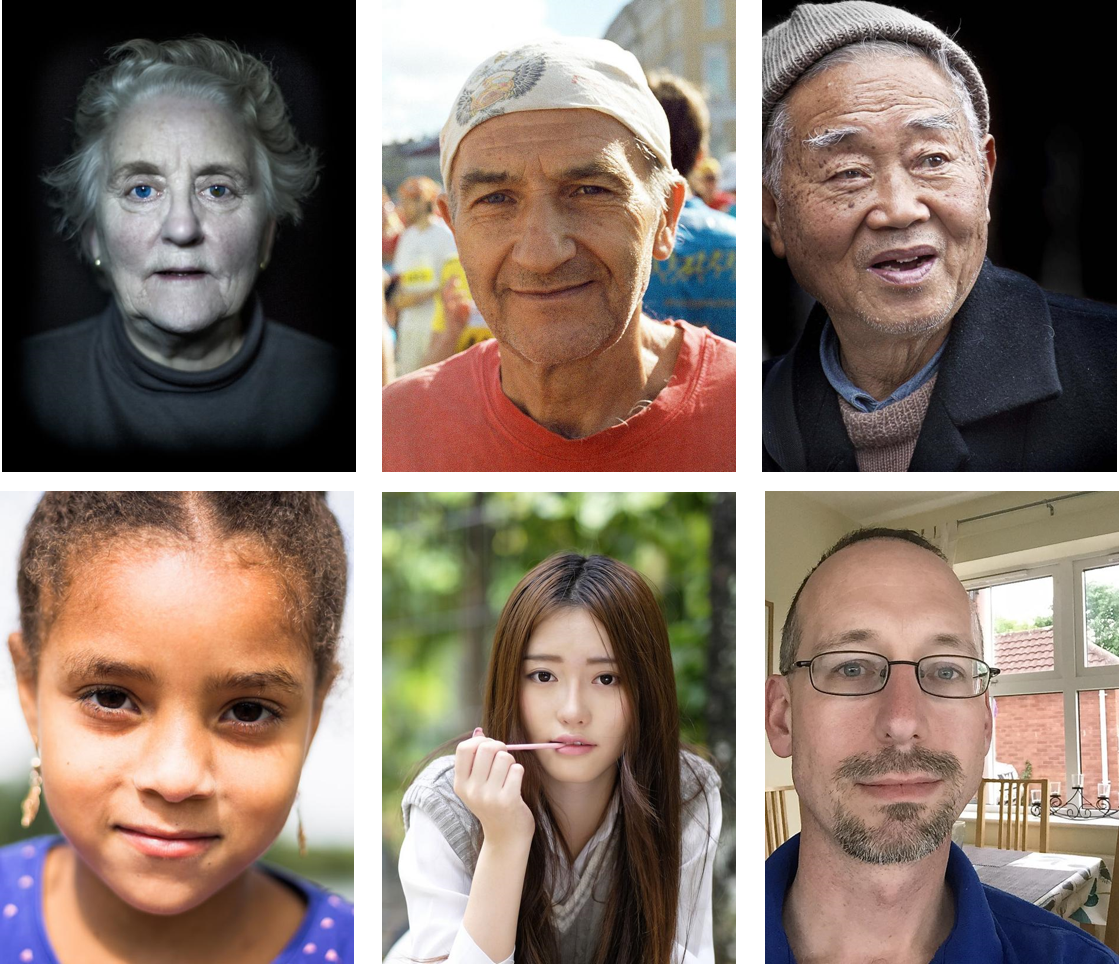}
\end{center}
   \caption{Example images of age group. Row1 : seniors, Row 2: child, youth and middle-aged person from left to right }
\label{fig:att_ex}
\end{figure}

\subsection{Analysis of the EG1800 dataset}
\label{sec:bias}


The public dataset EG1800~\cite{shen2016automatic} 
\Lars{contains} 1,800 images 
for portrait stylizing from segmentation results. 
\Lars{However, many urls are broken and the total amount of available images is 1,579.}
\Lars{Six human annotators \hj{\Lars{labeled} the attributes of each portrait image } with respect to race, gender, and age, and for images containing more than one person the annotators counted all of them.}
We divided \Lars{the} dataset \Lars{into} 18 classes
as shown in Table \ref{tab:att_label}, and 
\Lars{illustrated the dataset} bias 
in Figure \ref{fig:att_histo}.
Both train\Lars{ing} and validation sets have a severe bias to\Lars{wards} Caucasian and youth \Lars{and} middle-aged people, and we could not find any image \Lars{containing} a senior black woman.
\Lars{Detailed results can be found in the \hj{Supplemental materials.}}

We conducted a comparative experiment to understand how data augmentation methods can \yj{re}solve the data bias problems.
In Table~\ref{tab:att_acc}, Experiment 1 and 2 \yj{use} the same model structure \yj{(ExtremeC3Net} \yj{trained with EG1800 dataset.}
However, they were trained with different data augmentation method.
\yj{For the Experiment 1}, we used naive data augmentation methods such as random resizing, random crop, and random horizontal flip.
\yj{For the Experiment 2}, we applied a more sophisticated data augmentation method proposed by PortraitNet,
\yj{which consisting of the deformation and the texture augmentation methods~\cite{zhang2019portraitnet}.}



\Lars{For the gender and age attributes, the number of images in the dataset has an impact on accuracy.}
The number of female images is 280 more than the number of male images in the training set, and \yj{it seems to make the model over-fitted} to the female portrait images.
The numbers of child and seniors images \yj{were} lower than youth \Lars{images}, but the accuracy \Lars{for child images was still high compared to that of youth images. }
\Lars{On the} contrary, the accuracy of seniors is remarkably lower than the others.
\yj{From the results,} we conjecture that the bias between the races is not that important compared to those of other attributes, but the bias from the gender and age makes meaningful accuracy degradation. 
\yj{Also}, the number of images for the child group \yj{did not bring a significant} imbalanced impact on accuracy, \yj{compared to the senior group case.}

The sophisticated data augmentation method \yj{used in Experiment 2 was effective for improving} the accuracy, \yj{but it could not solve the imbalance problem} completely\yj{;} the accuracy of seniors was still lower than \yj{the} other attributes.
\yj{The possible reason of the phenomenon would be} that the distinct features of the seniors \yj{in Figure~\ref{fig:att_ex}}, such as wrinkles and ages spots, \yj{would be} difficult to \yj{be covered} by the data augmentation method.
From the results, \yj{the augmented data from the proposed data generation framework was shown to be effective} for all the attributes and improved the balanced accuracy of each attribute.
The accuracy disparity in an the age groups reduced from $1.82$ (Experiment 2) to $1.12$ (ours).



%% file: Eng/Conclusion.tex
\section{Conclusion}
In this paper, we proposed \hj{ExtremeC3Net which is an} extremely lightweight two-branched model consisting of CoarseNet and FineNet \yj{for solving the portrait segmentation task.} 
The coarseNet produces \yj{the} coarse segmentation map and \yj{the} FineNet assists \yj{spatial details to catch the boundary information of the object}.
\yj{We also proposed the advanced C3-module for each network which elaborately adjusts the dilation ratio of the convolutional filters according to their positions.} 
\yj{From the experiments on a public portrait segmentation dataset,} our model \yj{obtained} outstanding performance \yj{compared to the} existing lightweight segmentation models. 
Also, we proposed a simple data generation framework \yj{covering the} two situation\yj{s}: 1) having human segmentation ground truths 2) having only raw images. 
\yj{We also analysed the configuration of the public dataset given portrait attributes and studied the effects of the bias among each attribute regarding the accuracy.}
The \yj{additionally labeled samples} \yj{we generated} \yj{were shown to be helpful for improving the} segmentation accuracy for all \yj{the attributes.}
\label{sec:endLOL}

%% file: SuppleEng/intro.tex
\section*{Appendix}
\label{intro_supp}
In this supplementary material, we provide additional results and methods that we could not include due to space limitation.
We indicate detailed number of dataset histogram from our additional annotations.
We also explain the method for measuring floating-point operations (FLOPs) and show other segmentation examples for qualitative comparison of our method with two extreme lightweight segmentation models based on ContextNet\cite{poudel2018contextnet} and ESPNet\cite{mehta2018espnet}.

%% file: SuppleEng/histo.tex
\subsection*{Detailed results of additional annotations}
\label{flop}


\begin{table*}[h]
\begin{center}
 
    \begin{tabular}{|ccc|ccc|}
    \hline
    \multicolumn{3}{|c|}{Training set } & \multicolumn{3}{c|}{Validation set} \\
    \hline
    Class & {\# Frequency} & {Cumulative value (\%)}  & Class & {\# Frequency} & {Cumulative value (\%)} \\
    \hline \hline
     {4} & 575   & 3.81\% &  {4} & 93    & 34.32\% \\
     {1} & 326   & 28.66\% &  {1} & 77    & 62.73\% \\
     {10} & 79    & 32.93\% &  {3} & 26    & 72.32\% \\
     {7} & 59    & 36.97\% &  {10} & 24    & 81.18\% \\
     {2} & 56    & 80.79\% &  {7} & 17    & 87.45\% \\
     {3} & 53    & 82.77\% &  {0} & 10    & 91.14\% \\
     {0} & 50    & 83.16\% &  {16} & 7     & 93.73\% \\
     {16} & 38    & 87.65\% &  {5} & 5     & 95.57\% \\
     {5} & 26    & 88.03\% &  {2} & 3     & 96.68\% \\
     {9} & 15    & 89.18\% &  {9} & 3     & 97.79\% \\
     {13} & 12    & 95.20\% &  {6} & 2     & 98.52\% \\
     {6} & 5     & 95.58\% &  {14} & 2     & 99.26\% \\
     {8} & 5     & 95.73\% &  {13} & 1     & 99.63\% \\
     {11} & 5     & 96.65\% &  {15} & 1     & 100.00\% \\
     {15} & 5     & 96.72\% &  {8} & 0     & 100.00\% \\
     {12} & 2     & 97.10\% &  {11} & 0     & 100.00\% \\
     {14} & 1     & 100.00\% &  {12} & 0     & 100.00\% \\
     {17} & 0     & 100.00\% &  {17} & 0     & 100.00\% \\
     \hline
    Total   & 1312  & 100.00\% & Total   & 271   & 100.00\% \\
\hline
  \end{tabular}%
  \end{center}
   \caption{Detailed number of dataset histogram of each group image}
  \label{tab:histo_num}%
\end{table*}%

Table~\ref{tab:histo_num} is detailed numbers of dataset histogram figure in Experiment section. 
Most of both training and validation sets consist of Caucasian and Youth or Middle-aged person, as shown in Table~\ref{tab:histo_num}.
Even, some of the group images do not exist in the dataset, and the bias of configuration in the dataset is more severe in validation set such as senior Black woman, Black child, and senior Asian man.
The total frequency is greater than the total number of data sets (training: 1,309 and validation: 270) because sometimes there is more than one person in the image.

%% file: SuppleEng/img.tex
\begin{figure*}[h]
\begin{center}
\begin{tabular}{c}
    \includegraphics[width=0.90\linewidth]{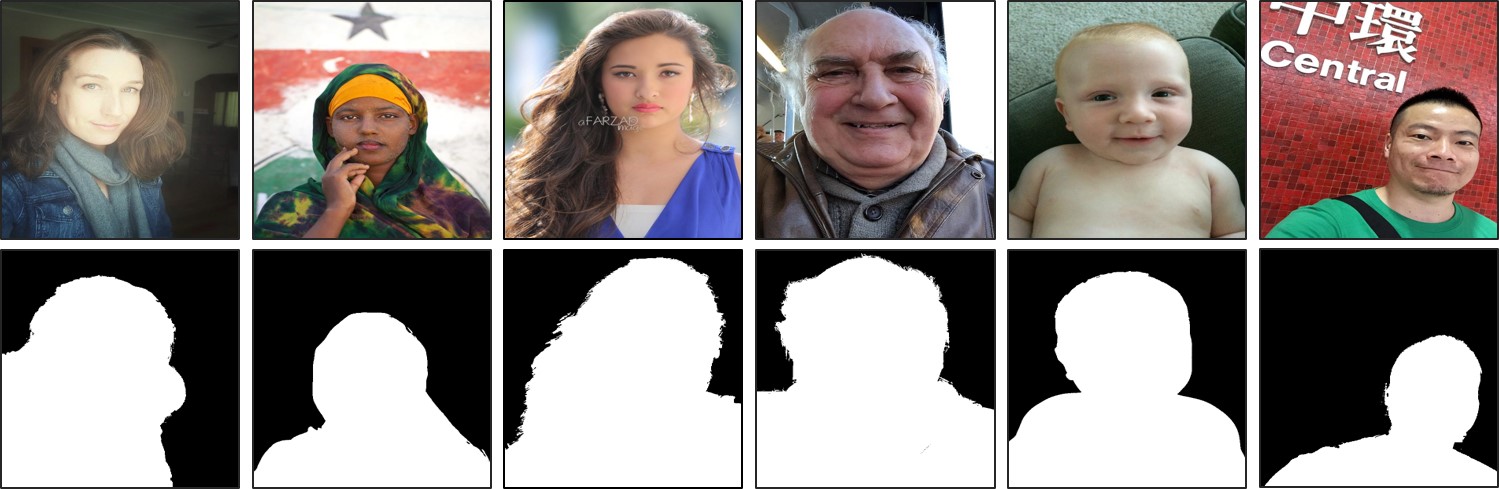} \\ 
    (a) Input images and ground-truths \\
    
     \includegraphics[width=0.90\linewidth]{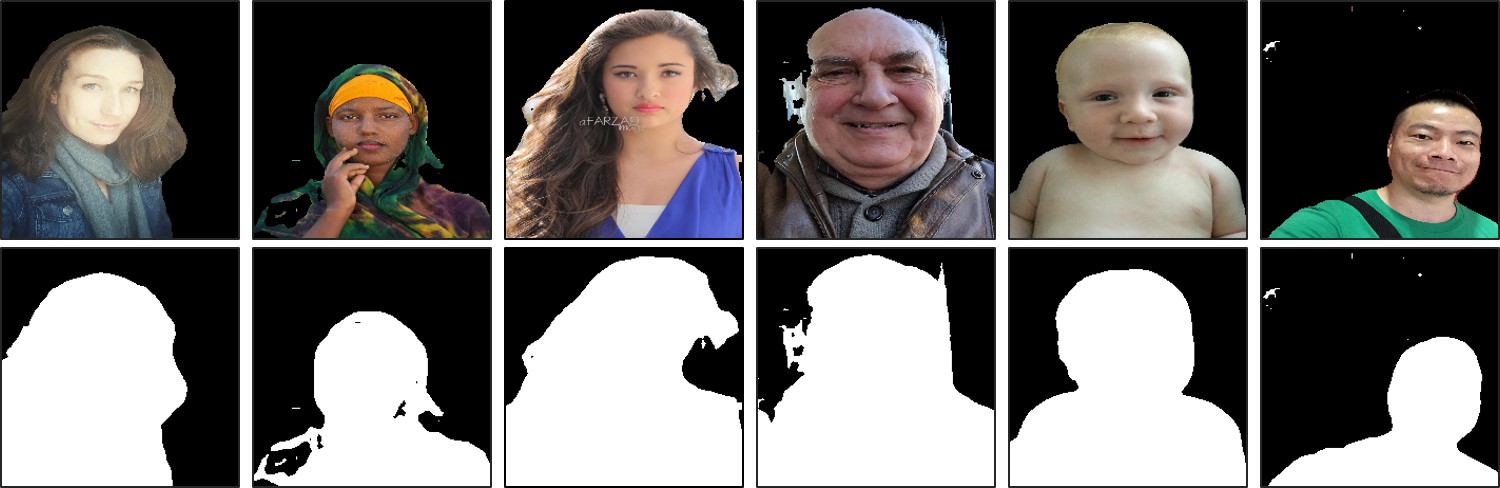} \\
     \textbf{(b) Ours ExtremeC3Net param : 37.9 K, mIoU : 94.98 }  \\
     
     \includegraphics[width=0.90\linewidth]{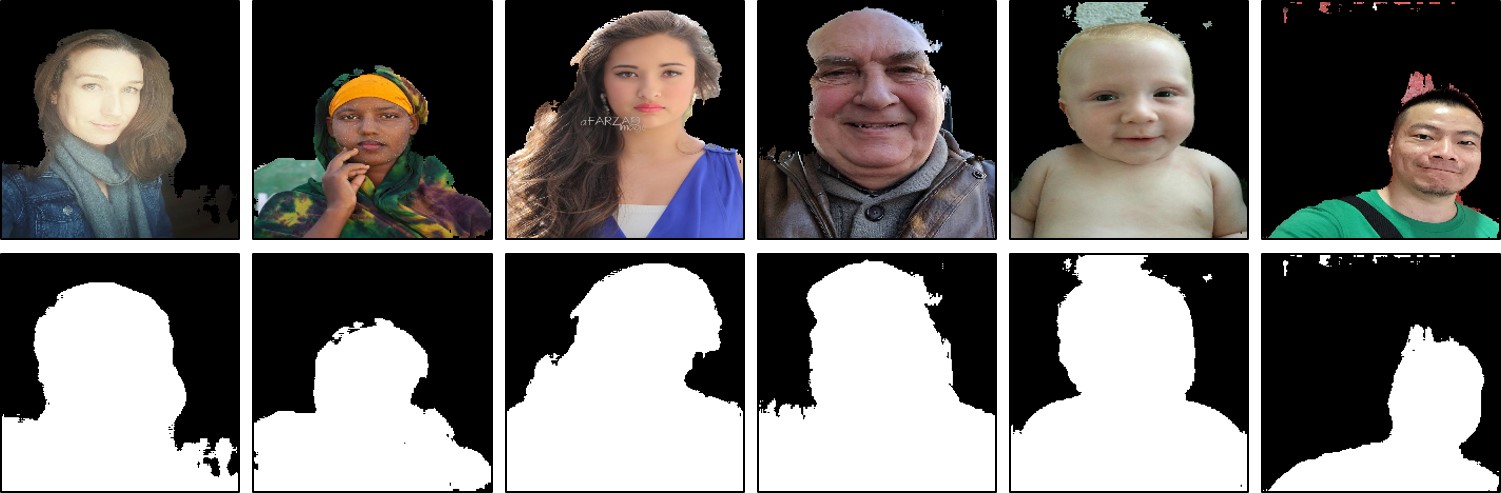}\\
     (c) DS-ESPNet(0.5) Param : 63.9 K, mIoU : 92.59 \\
     
      \includegraphics[width=0.90\linewidth]{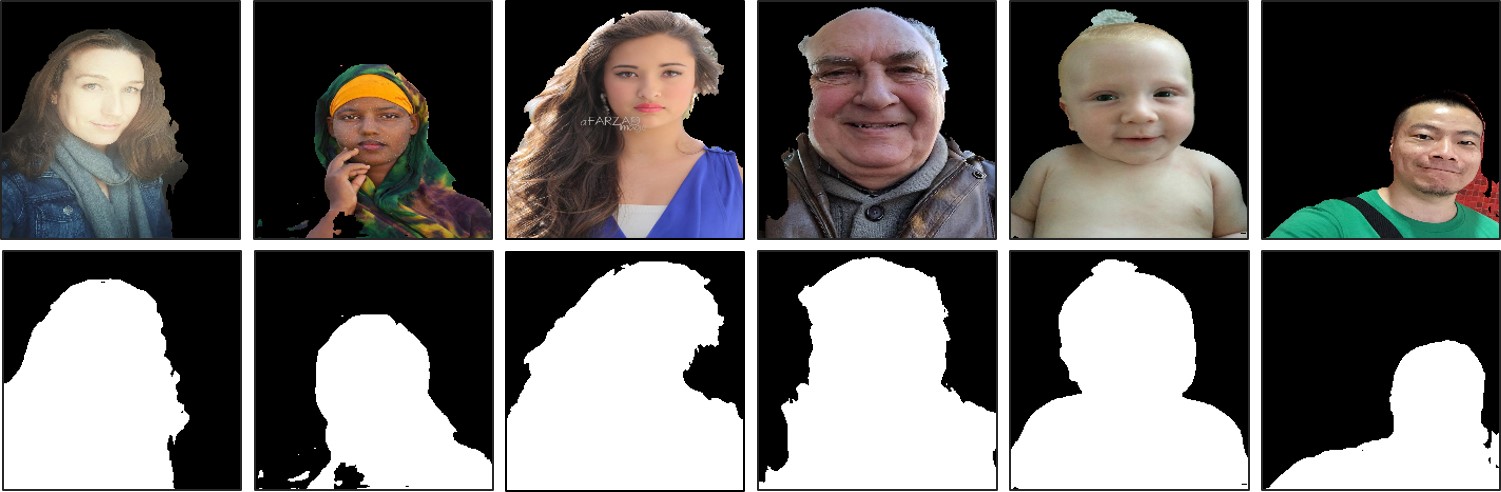} \\
       (d) ContextNet12(0.25) Param : 67.2 K, mIoU : 93.24 \\
        \end{tabular}%
\end{center}
   \caption{Additional qualitative comparison results on the EG1800~\cite{shen2016automatic} validation dataset.}
\label{fig:ex}
\end{figure*}

\subsection*{Additional qualitative comparison results}
\label{img}

We show other example results with covering various attributes that we could not illustrate in the Experiment section due to space limitation.
Our model has less complexity than other extreme lightweight model but shows better performance as shown in Figure~\ref{fig:ex}.

%% file: SuppleEng/flop.tex
\subsection*{FLOPs Calculation}
\label{flop}

\begin{table*}[h]
\centering
    \begin{tabular*}{0.99\textwidth}{@{\extracolsep{\fill}} ccc}
    \centering{Layer} & Operation & Flop \\
     \hline \hline
    {Convolution}& $O=F*K$  & $2\cdot H_oW_o\cdot K_hK_w\cdot C_iC_o/g$ \\
     \hline
  {Deconvolution}& $O=F*K$  &  $2\cdot H_iW_i\cdot K_hK_w\cdot C_iC_o/g$ \\
     \hline
     {Average Pooling}& $O=Avg(F)$ & $H_i\cdot W_i\cdot C_i$ \\
     \hline
    Bilinear upsampling & $f(x,y)=\sum_{i=0}^{1}\sum_{j=0}^{1}a_{ij}x^iy^j$ 
    & $3\cdot H_i\cdot W_i\cdot C_i$  \\
     \hline
      {Batch normalization}& $(F-mean)/std$ &  $ 2\cdot H_i\cdot W_i\cdot C_i$ \\
     \hline
    {ReLU or PReLU}& $O=g(F)$  & $ H_i\cdot W_i\cdot C_i$ \\
     \hline
 
    \end{tabular*}%

      \caption{The detail method for calculating FLOPs}
  \label{tab:flop}%
\end{table*}%
Table \ref{tab:flop} shows how we calculated FLOPs for each operation.
The following notations are used.  \newline
$F$ : A input feature map \newline
$O$ : A output feature map \newline
$K$ : A convolution kernel \newline
$K_h$ : A height of convolution kernel \newline
$K_w$ : A width of convolution kernel \newline
$H_i$ : A height of input feature map \newline  
$W_i$ : A width of input feature map \newline 
$C_i$ : A input channel dimension of feature map or kernel \newline
$C_o$ : A output channel dimension of feature map or kernel \newline
$g$ : A group size for channel dimension \newline
$H_o$ : A height of output feature map \newline
$W_o$ : A width of output feature map \newline
$g(\cdot)$ : A non-linear activation function\newline